\documentclass[english]{iithesis}

\usepackage[utf8]{inputenc}

\polishtitle {Zastosowanie metod uczenia nadzorowanego i ze wzmocnieniem do tworzenia agentów opartych na sieciach neuronowych do gry StarCraft II}
\englishtitle   {Applying supervised and reinforcement learning methods to create neural-network-based agents for playing StarCraft II}

\englishabstract{Recently, multiple approaches for creating agents for playing various complex real-time computer games such as StarCraft II or Dota 2 were proposed, however, they either embed a significant amount of expert knowledge into the agent or use a prohibitively large for most researchers amount of computational resources. We propose a neural network architecture for playing the full two-player match of StarCraft II trained with general-purpose supervised and reinforcement learning, that can be trained on a single consumer-grade PC with a single GPU. We also show that our implementation achieves a non-trivial performance when compared to the in-game scripted bots. We make no simplifying assumptions about the game except for playing on a single chosen map, and we use very little expert knowledge. In principle, our approach can be applied to any RTS game with small modifications. While our results are far behind the state-of-the-art large-scale approaches in terms of the final performance, we believe our work can serve as a solid baseline for other small-scale experiments.}

\polishabstract{W ostatnim czasie zaproponowano wiele podejść do tworzenia agentów do grania w różne złożone gry komputerowe czasu rzeczywistego, takie jak StarCraft II czy Dota 2, jednak albo zawierają one w agencie znaczną ilość wiedzy eksperckiej, albo wykorzystują zbyt duże dla większości badaczy zasoby obliczeniowe. Proponujemy architekturę sieci neuronowej do rozgrywania pełnego dwuosobowego meczu w StarCraft II, trenowaną za pomocą uczenia nadzorowanego i ze wzmocnieniem ogólnego przeznaczenia, która może być trenowana na pojedynczym PC z pojedynczą jednostką GPU. Pokazujemy również, że nasza implementacja osiąga nietrywialną skuteczność w porównaniu do oskryptowanych botów w grze. Nie przyjmujemy żadnych upraszczających założeń dotyczących gry, z wyjątkiem rozgrywki na pojedynczej wybranej mapie, i wykorzystujemy bardzo niewiele wiedzy eksperckiej. W teorii, nasze podejście z niewielkimi modyfikacjami może być zastosowane do dowolnej gry RTS. Chociaż nasze wyniki są daleko w tyle za najlepszymi podejściami na dużą skalę pod względem ostatecznej wydajności, wierzymy, że nasza praca może służyć jako podstawa dla innych eksperymentów na małą skalę.}

\author         {Michał Opanowicz \fmlinebreak \small{michal.opanowicz@gmail.com}}
\advisor        {dr Paweł Rychlikowski}
\date          {September 26, 2021}                   

\usepackage{tabularx}
\usepackage{graphicx}
\usepackage{amsfonts}
\usepackage{amsmath}

\begin{document}

\chapter{Introduction}

\section {AI in games}

Playing popular games has been a benchmark for artificial intelligence since the 1970s, with computers being used to play board games like chess or checkers. With the increase of computing power, search-based techniques became viable and humans were defeated in many board games, such as mentioned chess, checkers, and Reversi.

Playing complex real-time games such as StarCraft II or Dota 2 however, remained largely out of reach for a long time. Search-based approaches are often inapplicable to them due to the extremely large state and action spaces, and because of that methods based purely on expert knowledge were usually used. Due to the complexity of those games however, hand-crafting a policy capable of acting in every situation is effectively impossible. That means those agents are usually easily exploited by human players.

In recent years with the development of neural networks and deep learning, especially deep reinforcement learning, new possibilities have opened. Using Convolutional Neural Networks combined with Monte Carlo Tree Search playing board games with relatively large action spaces such as Go became possible \cite{alphago}. At the same time, in many simple real-time games with small action spaces a superhuman performance was achieved through the development of approaches such as Deep Q-learning \cite{mnih2013playing}.

In 2018 OpenAI has presented a large-scale RL approach capable of playing Dota 2 that was able to defeat professional players repeatedly, known as OpenAI Five \cite{openaifive}. Not too long after that in 2019, DeepMind has presented a somewhat similar approach for playing StarCraft II \cite{alphastar}, which shows reinforcement learning is capable of achieving very strong performance. Unfortunately, those approaches usually require hundreds or thousands of years of in-game experience to train successful agents - learning to play games with the speed of a human remains largely unachievable.

\section{StarCraft II}

StarCraft II (SC2) is a complex, single- and multiplayer real-time strategy (RTS) game where players need to manage resources, build and command multiple units. It poses significant challenges for machine learning: the game engine makes 22 steps per second, and games tend to last more than 10 minutes and sometimes over an hour, meaning that an agent needs to deal with long time horizons; action space is also very large as the players command units by selecting points on the screen.

A brief description of the 1vs1 game in SC2:
\begin{enumerate}
    \item Each of the players chooses one of the three races they will play - Terran, Protoss, or Zerg. Each race has access to completely different units and some unique game mechanics, which has a significant impact on the strategy.  
    \item The players start the game with one base each, usually located at the opposite ends of the map, and the same amount of workers that collect resources and build production buildings or more bases. Resources in each base are limited, so creating more bases in longer games is necessary.
    \item The players build production buildings and offensive units and attempt to attack each other. Players can give commands to the units by clicking specific units or groups of them and selecting points on the map where the units should move or enemy units to attack. They can also use various abilities that the units might have, such as disabling opponent's weapons, healing, or increasing the armor of the allies in a certain radius.
    \item The game ends when one of the players loses all the buildings or leaves the game, in which case the other player wins. The game also ends in a draw when neither of the players builds or destroys any structure or collects resources for around 6 minutes.
\end{enumerate}

It should be also noted that enemy units are invisible (covered by the so-called fog-of-war) unless in the vision range of the allied units, which significantly complicates the strategies and enables various mind-games such as faking committing into some strategy, or denying scouting information.

For use in machine learning, DeepMind in cooperation with Blizzard Entertainment have developed StarCraft 2 Learning Environment (SC2LE, with main python library called PySC2) \cite{sc2env}, that allows computer programs to play the game using three different interfaces: 
\begin{enumerate}
    \item A true 3D rendering that is very similar to what the human players see (known as the \textit{render interface}),
    \item A 2D top-down view split into multiple feature layers that show various properties of the units and maps (known as the \textit{feature interface}),
    \item A list of units containing all of the information about them that is visible for the player (known as the \textit{raw interface}).
\end{enumerate} 
For sending commands, interfaces 1 and 2 allow the program to only send commands that would also be available to the human player, interface 3 however allows for much more powerful commands, such as selecting any desired subset of units located anywhere on the map. 

The game also includes different built-in scripted bots, ordered by their increasing difficulty: Very Easy, Easy, Medium, Medium-hard, Hard, Very Hard, Elite (names as listed in the PySC2 API). There are also Cheater 1, Cheater 2 and Cheater 3 bots which are stronger than the Elite bot and get various advantages (such as double the income, no fog-of-war), that would violate the rules of the normal game.

'Very Easy' bot avoids building large armies and can be defeated by a human without much experience in the game, 'Elite' requires a significant amount of skill, and some players estimate it plays on a Silver or Gold league level, so in the top 60-70\% of the players playing regularly. 'Medium' bot is fairly difficult for new players, and usually requires several hours of experience in RTS games to defeat it.

\begin{figure}
    \centering
    \includegraphics[width=12cm]{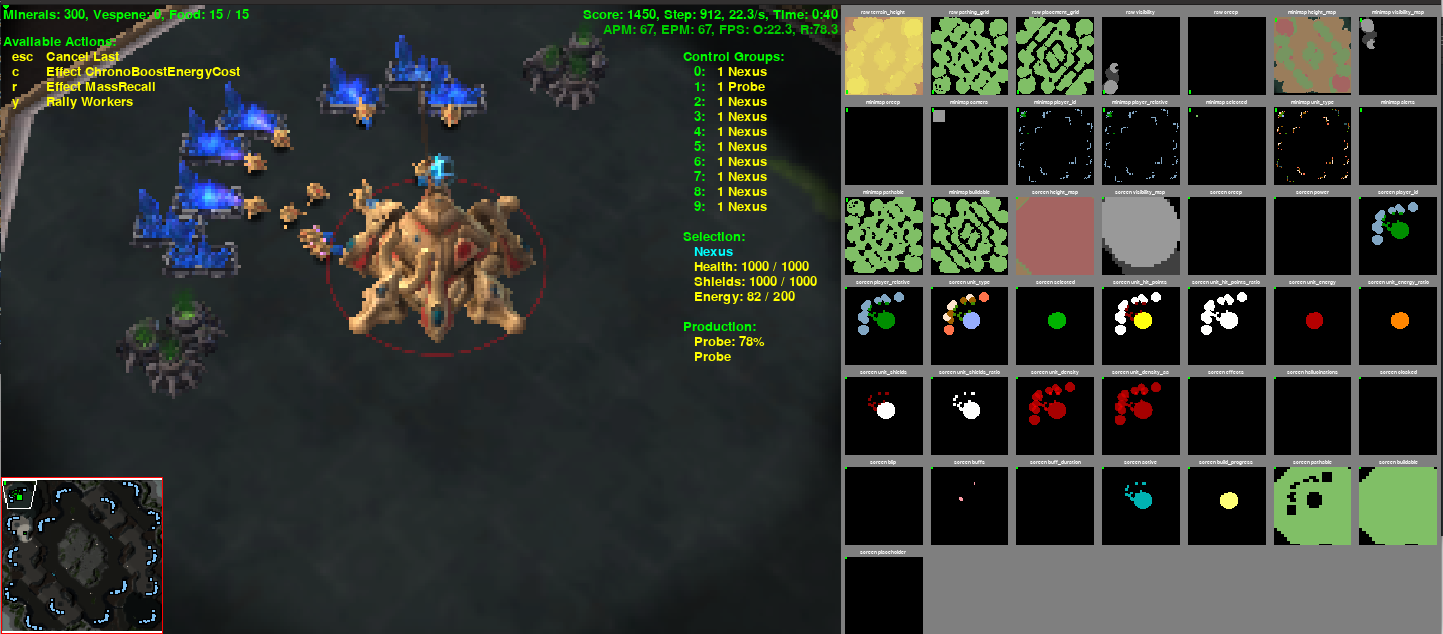}
    \caption{Visualisation generated by PySC2: rendering from the \textit{render interface} (left) next to the feature layers (with colors for human readability, right) from the \textit{feature interface}}
    \label{fig:my_label}
\end{figure}

\section{Previous work}

Since PySC2 was released in 2017 \cite{sc2env}, not many machine learning approaches attempting to play the full SC2 game using it were published. First approaches focused on solving 'minigames' - a set of custom maps, each focused on solving a specific objective - moving units around the map, fighting a group of enemies in a specific fixed situation, or building units with a limited action set. Playing the full multiplayer game of StarCraft II was mostly attempted using partially-scripted approaches with large parts of strategy built into the agent by experts. It should be noted that those agents can be quite good, for example TStarBot \cite{tstarbot} that was able to defeat all built-in AIs. Hybrid approaches also were proposed, with the agent architecture consisting of both hard-coded and learning components, which were successful at defeating bots in restricted conditions \cite{modular} \cite{reinforcement}. 

DeepMind in 2020 showed a powerful agent called AlphaStar \cite{alphastar}, that was controlled by a neural network trained on human replays and through self-play with reinforcement learning and played on a level comparable to human experts - clearly showing that using enough computational resources it is possible to play the game through pure machine learning. However, their implementation wasn't made public, and computational resources used for this experiment would be prohibitively expensive for most amateur researchers and small-scale labs. Two approaches based on AlphaStar were also published by other researchers \cite{scc} \cite{tstarbotx}. They used less resources than the original AlphaStar, but they limited their agents to playing only mirror matchups (Terran vs Terran or Zerg vs Zerg).

Recently, a small-scale approach that was based purely on imitation learning was published on GitHub \cite{sc2im}. The authors report it took a week to train using 4 GPUs and is capable of producing an agent that can defeat some of the built-in AIs. 

\section{Our contribution}

We combine recent developments in applying machine learning to the RTS games to produce an agent that achieves non-trivial performance in StarCraft II using low computational resources - specifically, around 10 days on a single PC with one GPU. We also show that with supervised pretraining, it is possible to obtain significant improvement using reinforcement learning even on our relatively small scale.

Our agent plays using a single recurrent neural network to process the information from the game and select an action. The network is trained to predict the probability distribution of the actions made by human players at every time step, known as behavioral cloning. The dataset consists of around 18000 games that were played on a single chosen map and include at least one Protoss player. Training takes around 7 days on a single NVidia GTX 1080 Ti GPU + 2 days to preprocess the replays. We adapt many parts of the AlphaStar architecture for our network, but we also use parts of the architecture described in Relational Reinforcement Learning \cite{rrl}. 

Our agents achieve higher win percentages against the built-in AIs than the ones presented in \cite{sc2im}. However, they are not directly comparable - agents from \cite{sc2im} were trained only on Terran vs Terran matches, while our agents play as a Protoss against all three races, but on a single map. 

Our implementation is written in Python, using PyTorch as a machine learning framework, and is available on GitHub: \href{https://github.com/MichalOp/StarTrain}{https://github.com/MichalOp/StarTrain}.

\chapter{Background}

\section{Neural networks}

Neural networks are a family of very powerful predictive models, used as classifiers or function approximators in various contexts. They usually have a form of a sequence or a directed graph of 'layers', where each layer is some parametrized linear transformation applied on a vector and an element-wise nonlinearity applied on the result. They are most commonly trained with gradient descent (as it is possible to efficiently compute the gradients using the chain-rule, such computation is usually called backpropagation), which makes it possible to use a lot of different loss functions.

In this section, we briefly describe the neural network components that we use in this work. 

\begin{subsection}{Multi-Layer Perceptron (MLP)}
    MLP module is a chain of so called fully-connected layers, each fully connected layer has a form:
    \[ \mathbf{y} = f(W\mathbf{x} + \mathbf{b}) \]
    where $W$ (called weight matrix) is a matrix of shape $m \times n$, $\mathbf{b}$ (called bias) is a vector of length $m$, $\mathbf{x}$ is the input vector of length $n$, $\mathbf{y}$ is an output vector of length $m$, and $f$ is a (usually element-wise) function called activation function, in modern neural network often the so-called ReLU (Rectified Linear Unit): $f(x) = max(x, 0)$. $W,b$ are the trainable parameters of the layer.
    
    In this work, we use ELU (Exponential Linear Unit) (as proposed in \cite{elu}) activation function:
    \[ ELU(x) = \begin{cases} x \text{ for } x > 0 \\ e^{x-1} \text{ for } x \leq 0   \end{cases} \]
    
    The authors of ELU show that it seems to allow for faster training and better generalization. In our case, it appears to improve the stability of training when operating on half-precision (16-bit) floating point numbers.
    
\end{subsection}

\begin{subsection}{Convolutional Neural Network (CNN)}
    
    CNNs are used to process inputs in which spatial relationships are important, such as images. They consist of convolutional layers which can be thought of as trainable filters - moving a small window over an image, and repeatedly applying a shared, small fully-connected layer to all pixels in the window, to produce a new, transformed 'image' (called \textit{feature map}), potentially with a different number of channels.
    
    \begin{figure}[t]
\includegraphics[width=8cm]{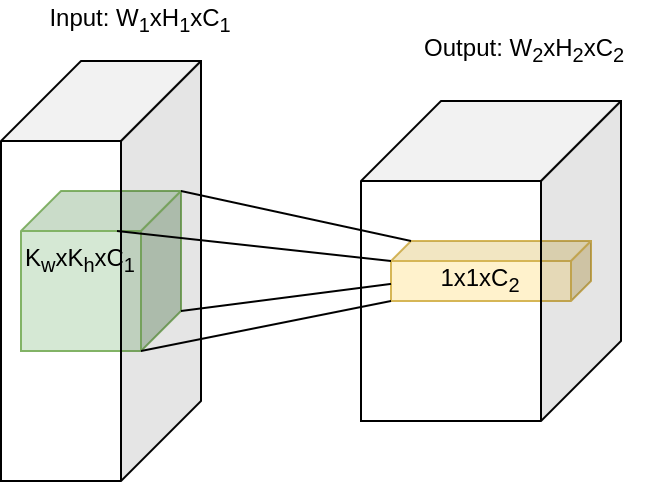}
\centering
\caption{Convolutional layer with a kernel in shape $K_w \times K_h$ and $C_2$ output channels is being applied to a 3-dimensional input tensor, producing a tensor in shape $W_2 \times H_2 \times C_2$, where $W_2 = W_1 - K_w + 1$, $H_2 = H_1 - K_h + 1$.}
\end{figure}

    A convolutional layer has 4 important hyperparameters: 
    \begin{itemize}
        \item the kernel shape, which is the width and height of the window that sweeps over the input,
        \item the number of input channels,
        \item the number of output channels,
        \item the stride, which describes how the window moves over the input - with stride=1, the window will be applied at every valid location, with stride=2, the window will only be applied at locations with coordinates divisible by 2, halving the width and height of the output image. 
    \end{itemize}
    
    Often, CNNs consist of interleaving layers with stride 1 that keep the number of channels, and layers with stride 2 that double it, applied until the output is an 'image' that has a small number of pixels, but a very large number of channels. 
    
\end{subsection}

\begin{subsection}{Residual Neural Network (ResNet)} 
    Residual neural networks were developed to allow training networks with a very large number of layers. Such networks are usually built from the following modules:
    \[ \mathbf{y} = f(\mathbf{x}) + \mathbf{x} \]
    where $\mathbf{x}$ is the input vector, $\mathbf{y}$ is an output vector, $f$ is some neural network with input and output that have the same shape, and the $ + \mathbf{x}$ term is a so-called residual connection, which allows the deeper layers to have the direct access to the input of earlier layers. 
    Residual neural networks with convolutional layers are currently considered the state-of-the-art on image classification tasks \cite{resnet}.
    
    In ResNets a so-called Batch Normalization \cite{batch} layer is often added to the $f$ neural network. This layer normalizes its inputs so that values have a mean 0 and variance 1 along the batch dimension and was shown to improve performance of deep ResNets. In our case, we replace Batch Normalization with Layer Normalization \cite{layer}, which normalizes the values along the layer dimensions, as we found it to improve the learning speed compared to Batch Normalization.
\end{subsection}

\begin{subsection}{Feature-wise Linear Modulation (FiLM)}
   FiLM \cite{film} is a relatively simple upgrade to the convolutional neural networks designed to `inject` some non-spatial information into the spatial input - originally introduced to create networks that process a piece of text and an image simultaneously, and answers the questions from the text by selecting points on the image. It is also useful in our case - when the non-spatial component of the network chooses to build a building, information about this can be introduced to the spatial component to choose the proper location on the screen where the building should be built.
   
   It works by using a small MLP applied to the non-spatial input to compute $a_c$ and $b_c$ coefficients for each channel of the spatial input, and then uses them to apply an affine transformation to the entire channel: 
    \[y[i,j,c] = a_c \cdot x[i,j,c] + b_c \]
   That way, the information gets 'evenly introduced' to the entire image. 
\end{subsection}

\begin{subsection}{Recurrent Neural Network (RNN)}
    Recurrent neural networks have an 'internal state' - an output that is passed as a part of the input in the next iteration. Such networks are usually trained by 'unrolling' them over several iterations, effectively creating a network consisting of a chain of modules with shared weights that can be trained as usual.
    
    Recurrent networks were originally used for the tasks such as natural language translation or sound processing, but they are also very useful in playing real-time games such as SC2 - they most importantly allow the network to remember or infer information about the game state that is not visible from a single observation.
    
    A simple RNN can look like this:
        \[ \mathbf{i}_{t} = concat(\mathbf{x}_{t}, \mathbf{y}_{t-1})\]
        \[ \mathbf{y}_{t} = f(W\mathbf{i}_{t}  + \mathbf{b}) \]
    which is similar to the fully-connected layer with the following differences: $W$ is a matrix of shape $m \times (n+m)$, $\mathbf{x}_{t}$ is the input vector for current iteration, $\mathbf{y}_{t-1}$ is an output vector from the previous iteration, and $\mathbf{y}_{t}$ is the output from the current iteration. 
    
    However, such RNNs are difficult to train, as they suffer heavily either from vanishing or exploding gradients. When computing the gradients, the transposed $W$ matrix will be applied to the back-propagated gradient many times, and if any of its eigenvalues is greater than 1, the gradients will start growing exponentially. To counteract that one can use an activation function that will limit the values, such as sigmoid, but that will on the other hand cause the gradients to vanish very quickly.
    
    To counteract that, an architecture called Long Short-Term Memory (LSTM) \cite{lstm} was introduced, in which one part of internal state is not multiplied by any matrix at all, and no nonlinearity is applied to it. The internal state can only be multiplied by factors smaller than 1, and a value smaller than 1 can be added to or subtracted from it. (The exact formulation is quite complicated so we recommend referring to the original paper for details.) This effectively removes the vanishing or exploding gradient problem, and allows for training on a very long sequences. In this work, we use LSTM as our RNN component.

\end{subsection}

\begin{subsection}{Transformer}
    Transformer \cite{trans} is an architecture introduced originally to process sentences in natural language (specifically for translation). It has been since then adapted to other tasks where a list of objects with some sort of relationship between them is present, often achieving state-of-the-art performance in such tasks. For example, Transformers have been used for image classification, achieving comparable results to ResNets \cite{visiontransformers}. 
    
    Transformer architecture is based on an attention mechanism, which intuitively allows a neural network to 'focus' on the part of some data that the network 'wants' to find at a given moment. This is done by computing a vector $q$ called 'query', and for every $i$th sample in the data vectors $k_i$ and $v_i$ called 'keys' and 'values'. Then, one can compute sample weights
     \[ s_i = q_i \cdot k_i \]
     \[ w_i = softmax(\mathbf{s})_i = \frac{e^{s_i}}{\sum_{j=1}^n e^{s_j}} \]
    and compute the output value:
     \[y = \sum_{i=0}^n w_iv_i\]
    In the case of Transformers, this step is usually done for each part of the data, which means that for example every token in the text 'asks' about all the other tokens.
    
    In our case, we use Transformers as proposed in Relational Deep Reinforcement Learning \cite{rrl}, where the Transformer module is applied to the 'flattened' output of the convolutional layer that looks on the screen - the idea is that this might allow the network to apply reasoning that requires understanding relationships between objects, such as 'this is an idle worker, so let's search for mineral fields in the vicinity for it to mine', or 'this is a combat unit, so lets search for enemy units that it counters and command it to attack', etc.

 \section {Reinforcement learning}
Reinforcement learning (RL) is a family of problems where a learning agent interacting with an environment receives a scalar reward for its actions, and its goal is to maximize said reward. Typically, a problem in reinforcement learning is described as Markov's Decision Process (MDP), which is a tuple $(S, A, P_a, R_a)$. $S$ is a set of states, $A$ is a set of actions the agent can make, $P_a(s_1,s_2)$ is a probability of transitioning from a state $s_1$ to a state $s_2$ when the agent selects an action $a$, and $R_a(s_1,s_2)$ is the reward that the agent gets in such transition.

An agent usually chooses actions using a policy $\pi$, where $\pi(s, a)$ is a probability of making an action $a$ in a state $s$. An agent playing in the environment generates a chain of observation-action-reward tuples, called a \textit{trajectory}: $(s_0,a_0,r_0),$ $(s_1,a_1,r_1),$ $(s_2,a_2,r_2),$ $..., $ usually ending in some terminal state.

To make optimization easier, a proxy objective is introduced - at any step $t$ in a state $s_t$, we want the agent to maximize the expected discounted sum of rewards $V_\pi(s_t) = \mathbb{E}_\pi \sum_{n=0}\gamma^n r_{t+n}$, where $\gamma$ is the discount factor $< 1$, usually $> 0.9$. We will call $V(s)$ the \textit{value} of a state. This objective has the nice property of being bounded if the rewards in MDP are bounded, and resembles a natural intuition that events in the near future are more important than events in the far future.

The goal becomes now to choose the $\pi$ in such a way that it maximizes $V_\pi$ for any state. This is however a quite difficult task, as in most cases the state space $S$ is gigantic. In deep reinforcement learning, the policy (represented by a neural network) is optimized locally, using Monte Carlo estimates of the true $V_\pi$. In this work, we will focus on methods based on the \textit{advantage} of an action - the difference between the value in a state $s$ and the value under an assumption a specific action is chosen in the state $s$. To compute the advantage estimate efficiently, another part of the agent is introduced - an estimator of the true value $\tilde{V}$, also represented by a neural network.

To compute the very noisy Monte Carlo estimate of the $V_\pi$ one can use an equation
\[ \tilde{V}_{MC}(s_t) = \sum_{n=0}^{k-t}\gamma^n r_{t+n} \]

However, since computing this estimate requires acting in the environment until a terminal state is reached, which in case of SC2 would mean potentially thousands of steps, to allow for more frequent updates a bootstrapped estimate is used instead - a short trajectory of length $l$ is generated, and the estimates are computed as follows:
\[ \tilde{V}_{bootstrap}(s_t) = \sum_{n=0}^{l-t-2}\gamma^n r_{t+n} + \gamma^{l-t-1} \tilde{V}(s_{l})\]
Where  $\tilde{V}(s_{l})$ is the neural network value estimate of the last state in the trajectory, called \textit{bootstrap value}. 

The advantages are then computed as follows - first, a trajectory $(s_0,a_0,r_0),$ $(s_1,a_1,r_1),$ $ ...,$ $(s_k,a_k,r_k)$ is generated by running the current policy $\pi$ in the environment. Then, an advantage estimate $A$ can be computed:
    \[ A_{\pi}(s_t) = \tilde{V}_{bootstrap}(s_t) - \tilde{V}(s_t) \]
Since we know what action was chosen at the step $t$, $A_{\pi}(s_t)$ is a bootstraped estimate of the true advantage.

Having the advantage computed, it is possible to write an optimization objective for the policy. Several such objectives were introduced - for more information see papers on the Asynchronous Advantage Actor Critic \cite{a3c}, Trust Region Policy Optimization \cite{trpo}, Proximal Policy Optimization \cite{PPO}, IMPALA \cite{impala}.

We use the Proximal Policy Optimization objective, as it is simple to implement and was shown to be empirically quite stable in a variety of conditions. The optimization objective is defined as follows:
\[J_{policy} = A_{\pi_{old}}(s_t) clip \left(\frac {\pi(s_t, a_t)}{\pi_{old}(s_t, a_t)}, 1 - \epsilon, 1 + \epsilon \right) \]
where $\pi$ is the policy being currently optimized, and $\pi_{old}$ the policy that was used to compute the advantages. Detaching the two allows applying the policy update several times on the same trajectory, as long as the current policy is not too different from the old one - the tolerance for this difference is controlled by the $\epsilon$ hyperparameter, usually set to the value of $0.1 - 0.2$. We use a conservative $\epsilon = 0.1$.

The $\tilde{V}$ estimator is usually optimized by minimizing the mean squared error between the neural network estimate and the Monte Carlo estimate:
\[ J_{value} = \frac 1 2 \left(\sum_{n=0}^{k-t}\gamma^n r_{t+n} - \tilde{V}(s_t)\right)^2 \]

Due to all estimates being very noisy, the data is usually processed in large batches, so that the gradient descent updates remain relatively stable. In our context, we have found out batches of size 512 were necessary for the training to remain stable.
    
\end{subsection}

\chapter{Network architecture}

\section{Input}
The network receives input as tensors of different shapes that represent what a human would see on a screen in a more machine-readable format - \textit {feature interface} from the introduction. Those inputs provide information that is directly available to the human player, however, there are some notable differences - the game screen is shown for humans as a 3d RGB rendering, but for the agent as a 2d top-down view with 27 layers, with individual layers describing specific properties of the units in the field of view. This is done both to reduce the rendering time and to allow the agent to focus on understanding the game instead of computer vision. 

This is the main difference in the problem formulation between our work and AlphaStar. AlphaStar has used the \textit{raw interface} that gives the agent superhuman knowledge, as the agent sees all of its units for the entire time ('camera' in AlphaStar is merely a movable rectangle on the map where certain commands can be issued and all properties of enemy units are visible). It also gives the agent the potential to use superhuman capabilities, such as precisely moving units in any place on the map without moving the 'camera' (although AlphaStar did not seem to use those capabilities).

Description of all input sources for the network can be found in the Table \ref{table:inputs_table}. We have extended certain inputs beyond what was available in the environment by default in places where we felt the information available to the network does not resemble the information available to the human players well. 

\begin{table}[ht]
\begin{tabularx}{\textwidth}{|llX|}
\hline
Input name         & Input shape   & Description \\ \hline
feature\_screen    & (27, 64, 64)  & The agent's view of the game screen. \textbf{Extended with information about the point on the screen previously selected by the agent.}   \\ \hline
feature\_minimap   & (11, 64, 64)  & The agent's view of the game minimap. \textbf{Extended with information about the point on the minimap previously selected by the agent.}    \\ \hline
cargo              & (N, 7)        & Units in a currently selected transport vehicle. \\ \hline
control\_groups    & (10, 2)       & State of the 10 control groups that the player has - number of units in a control group and a type of unit that was first added to the control group.  \\ \hline
control\_groups\_hint    & (10, 2)   & State of the 10 control groups that the player we are trying to imitate had in the middle of the game - \textbf{added by us.}  \\ \hline
multi\_select      & (N, 7)        & Units selected when selecting multiple units.             \\ \hline
player             & (11)         & Information about various resources and scores that the player has - minerals, vespene, supply, game\_score, etc.            \\ \hline
production\_queue  & (N, 2)        & Units queued for production in a currently selected production building. \\ \hline
single\_select     & (N, 7)        & Information about the unit that is currently selected, if a single unit is selected.     \\ \hline
game\_loop         & (1)          & Game time, \textbf{normalized to reach 1 after 1 hour}.            \\ \hline
available\_actions & (573)        & Actions that are currently valid.           \\ \hline
prev\_action       & (1)          & Previous action the agent did, \textbf{modified to show the last meaningful action (not a no-op)}. \\ \hline
build\_order       & (20)         & First 20 units and structures that were constructed by the player we are trying to imitate, \textbf{added by us}.   \\ \hline
mmr                & (6)          & The MMR (Match Making Rating, the skill estimate) of the player we are trying to imitate, divided by 1000, one-hot, \textbf{added by us}. \\ \hline
\end{tabularx}
\caption{All of the neural network's inputs. N in shape denotes that the input has variable length. For such inputs up to 32 first entries are considered, the rest are ignored. Descriptions in \textbf{bold} denote our modifications of the base environment.}
\label{table:inputs_table}
\end{table}

\section{Spatial processing}

Probably the most important and complex parts of the agent's input are the spatial processing columns that use \textit{feature\_screen} and \textit{feature\_minimap} and extract information from them for action selection and selecting points on the screen. The general design follows the map processing module used in AlphaStar, however, there are some important distinctions, most notably the use of attention for relational processing of the features produced by the convolutional network as proposed in the Relational Deep Reinforcement Learning \cite{rrl}. 

The following steps are applied to process spatial observations:

\begin {enumerate}
\item Screen and minimap features might be either categorical (unit type, is the building under construction, etc.) or numeric (health, energy, etc). Binary categorical features are left as-it, non-binary are expanded using trainable embeddings (effectively equivalent to using one-hot representation followed by an MLP), and numeric features are scaled into the 0-1 range based on their maximum possible value.
\item After embedding and scaling, features are processed by 3 convolutional layers with stride 2, which change the size of the feature maps from 64x64x32 to 8x8x128. Between the application of the convolutional layers intermediate values are stored as \textit{screen\_bypass} and \textit{minimap\_bypass}, to preserve precise spatial information for selecting points on the screen.
\item 8x8 features are then processed by several residual blocks with FiLM, conditioned on the \textit{scalar\_input}, described in the next section.
\item Features from screen and minimap are then flattened along width and height and joined to form a single tensor in shape 128x128, and processed by a Transformer self-attention layer.
\item A shared LSTM is then applied to each of the items from the Transformer outputs, storing information about each pixel of the feature map separately. This is done to hopefully allow the network to remember some facts about the relatively precise state of the game map, potentially already including some results from relational reasoning performed by the Transformer layer.
\item LSTM output is then processed by two Transformer layers, to further apply relational reasoning to the entities on the map.
\item Processed features from the previous step are then separated and reshaped back into 8x8x128 feature maps, to be used in transposed convolutional layers for selecting points on the screen or minimap.
\end {enumerate}

\section{Non-spatial processing}

Similarly to spatial features, non-spatial ones can be categorical or numeric. We use the same technique to preprocess them as with the spatial features. 

For further processing, we use two types of modules - for features of constant length we use MLP, and for features of variable length (they are in essence lists of items) we first apply an MLP to each item, and then aggregate the list using MaxPool along the list dimension, to produce an output in the size of a single item - such approach was used in OpenAI Five \cite{openaifive}. This can be thought of as skimming over the entries and only remembering some things that stand out. 

Outputs of those layers are then concatenated into a single tensor called \\ \textit{scalar\_input}. We also concatenate selected outputs: \textit{multiple\_select}, \textit{single\_select}, \textit{available\_actions}, \textit{build\_order} to produce \textit{gating\_input}, that is later used during the action selection - intuitively, those are the inputs that should influence the selection of the action quite significantly, so adding an easier access for them should make training faster.

One other input that gets special treatment is the \textit{control\_groups} input - we found that controlling the control groups is very hard to learn for the network and also very influential, so we tried to improve the performance on it, by adding a bypass from the control groups input to the layers that make a selection using them. 

We also found that it was important to normalize the output vectors to have variance 1 and mean 0 before concatenating - presumably because different inputs might have different magnitudes despite earlier normalization.

After spatial input processing finishes, the \textit{scalar\_input} is concatenated with a flattened output of Transformer layers from spatial processing and passed through a 2-layer LSTM with 512 output size to produce \textit{lstm\_output}.

\section{Selecting actions}

Action space has a hierarchical structure - there are 573 actions that a player can do, however, each of those actions has also additional parameters - for example, action \textit{select\_screen}, which selects a unit at the specific point on the screen if there is any, requires coordinates on the screen and the information whether the clicked unit should be selected instead of the current selection, added to current selection or removed from it. 

To properly model probability distributions in this complex action space, we use an autoregressive model. We first sample the action, and then sample the required parameters one by one, from distributions conditioned on the selected action and previously sampled action parameters. A similar approach was also used in AlphaStar.

In more detail, it works as follows: first, one of the 573 possible game actions is sampled from the action distribution generated by MLP from \textit{lstm\_output}; then the result of that sampling is embedded into a vector of size of the \textit{lstm\_output} and added to \textit{lstm\_output}. After that, the \textit{delay} is sampled based on the new \textit{lstm\_output}, and again \textit{lstm\_output} is updated in the same fashion as before. The process is repeated until all required parameters are sampled. 

It should be noted that we only use this way of sampling for the \textit{action}, \textit{delay} and \textit{queued} outputs, as they are involved in most of the actions, and apart from them actions usually use only one or two other parameters. 

For sampling positions on the screen, the updated \textit{lstm\_output} is reshaped into size 8x8x16 and concatenated with the output of spatial input processing columns (that has shape 8x8x128), and later passed through 3 residual blocks with FiLM \cite{film}, also conditioned on \textit{lstm\_output}. Their result is subsequently combined with the \textit{screen\_bypass} and passed through transposed convolutional layers to produce 3 maps of logits of shape 64x64x1, from which \textit{screen1}, \textit{screen2} and \textit{minimap} outputs are sampled. Notably, selected points on the screen and the minimap are added to the agent's observation in the next step as an additional feature with one-hot encoding. 

During training, probability distributions are conditioned on the player's actions that occurred in the game that we are trying to imitate, and during play, they are conditioned on previous actions made by the agent.

Most of the non-spatial outputs use simple, 2-layer MLPs to select the action parameters, however the main action selection is done by a 4-layer MLP with residual connections, and its output is multiplied by an output of an additional MLP with sigmoid activations on the final layer that uses the \textit{gating\_input}. This mimics the similar part in AlphaStar, and aims to improve the quality of actions the network makes by providing it with as much information and processing capabilities as possible. Intuitively, the stack of 4 layers is supposed to choose the action the network wants to make, and the gating layer allows only the actions that make sense.

Other differently processed output is the set of outputs for the control groups. The control group id to act on is selected by a small attention layer that takes the control groups bypass computed earlier and the \textit{lstm\_output} as an input to make the selection possibly focused on the content of the control groups and independent of their order.   

\begin{figure}[t]
\centering
\includegraphics[width=15cm]{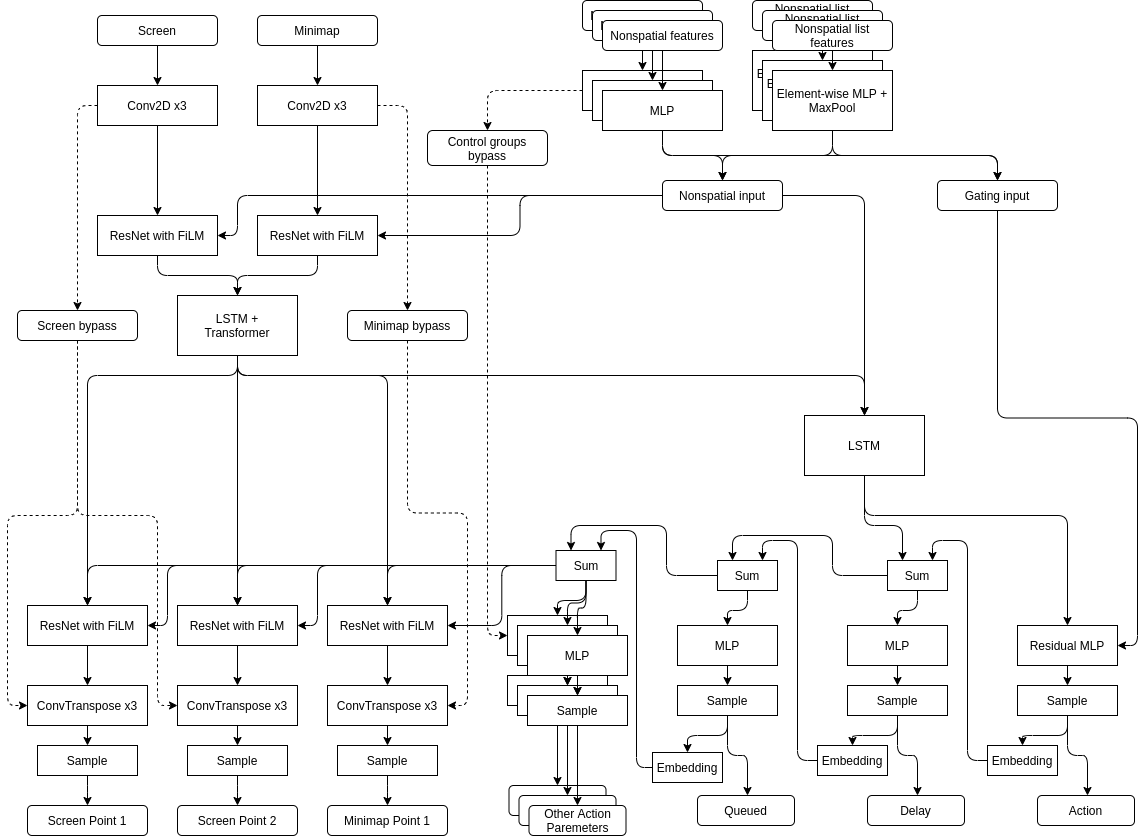}
\caption{A high-level diagram of the neural network. The boxes with sharp edges denote layers, the boxes with rounded edges denote values.}
\end{figure}

\chapter{Learning pipeline}

\section{Replay selection}

We use SC2 version 4.9.3 as this version has over 300000 1v1 replays available through Blizzard's API. From those replays we select ones that meet the following criteria:
\begin{itemize}
    \item At least one of the players is Protoss;
    \item At least one of the players is above 2500 MMR (roughly top 60\% of the leaderboard);
    \item The map the game is played on is 'Acropolis';
    \item The game is longer than 1 minute.
\end{itemize}

This gives us a dataset of around 18000 replays.

\section{Replay preprocessing}
SC2 replays aren't stored as a screen recording but as an entire list of actions that all players have made and the random seed, from which the entire game can be reconstructed. This means the game needs to be replayed to extract the observations for the training. To prevent unnecessary computation, we generate observations once and store them for later.

During the 'replaying' of the replays, we remove no-ops from the observations and add a 'delay' target for the network, similarly to AlphaStar. This means that in evaluation the network decides how long it will 'sleep' before the next action will be executed. We find this approach necessary as otherwise the sequences of observation-action pairs would mostly consist of very long chains of no-ops. We also remove chains of \textit{move\_camera} actions and store only the last position. This is done to circumvent the limitation of the game interface - human players can move their mouse to the edge of the screen to move the camera at a certain rate, but bots that play using PySC2 need to specify exact consecutive coordinates of the camera, so imitating this camera movement is quite difficult for the network.

At this stage, we also extract the \textit{build\_order} - first 20 units and buildings that the player that we are imitating has built, to condition the network on it. This step was also done for AlphaStar.

The preprocessing takes around 2 days on a 16-thread CPU (and is mostly CPU-bound), and the resulting dataset can be compressed to around 100 GB.
        
\section{Training}

\begin{subsection}{Supervised learning}
    We train our neural network to minimize cross-entropy loss between the actions done by the player and the probability distribution predicted by the network. We use an Adam optimizer with a learning rate equal to 0.001.

    Depending on memory constraints, the network can be trained using batches of size 32 or 16, and our implementation allows to accumulate gradients over many batches so that large batch sizes can be emulated using smaller true batch sizes for improved training stability on machines with small GPU memory size. We use a sequence length of 32 for training the recurrent parts of the network. 
    
    During supervised learning, we process the entire dataset around 7 times - this is nowhere near fitting the dataset, as seen on the loss plot \ref{loss_plot}. This suggests that using larger computational resources the network could be trained much longer for potentially significantly higher performance.
    
    \begin{figure}[ht]
    \centering
    \includegraphics[width=12cm]{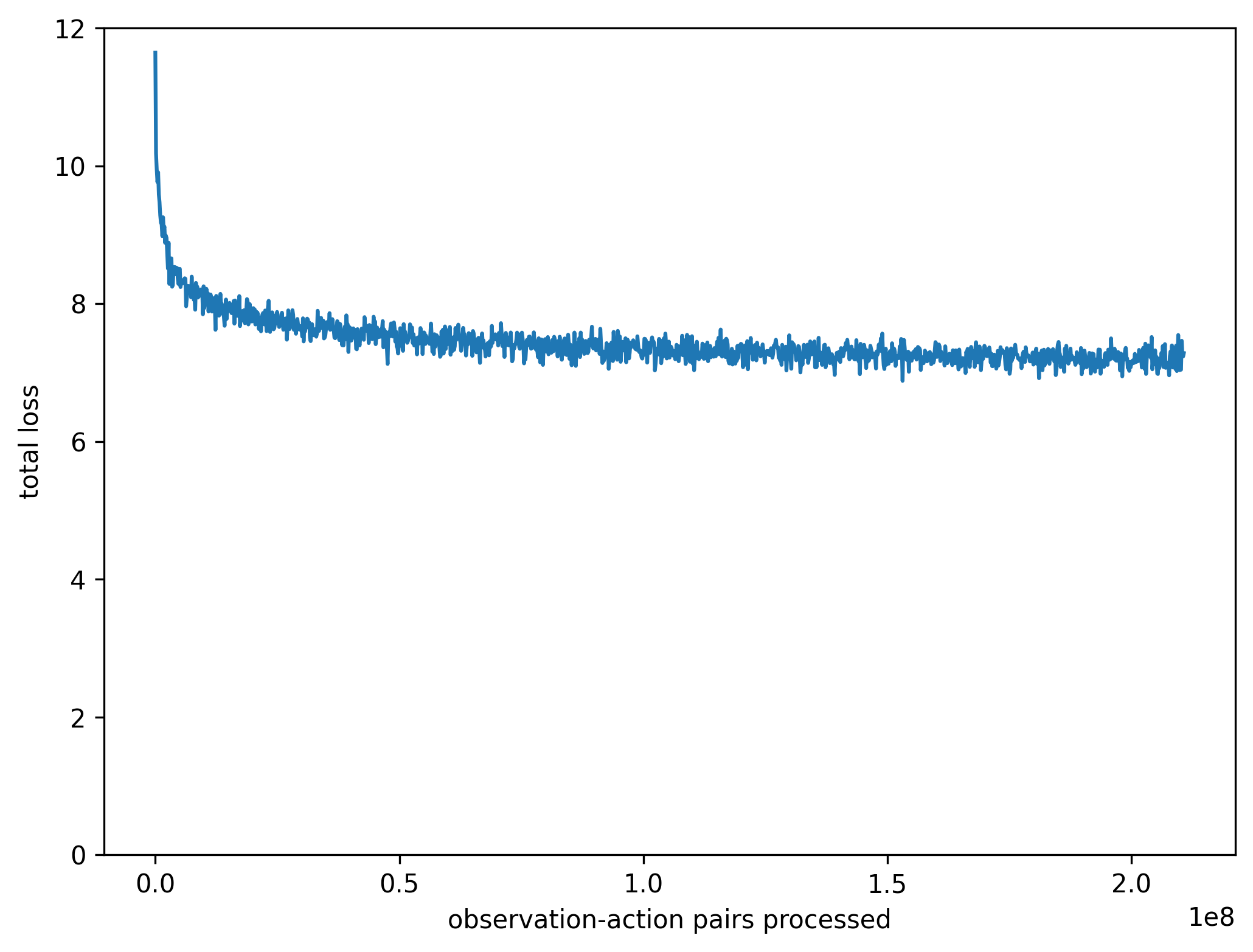}
    \caption{The average loss per sample during supervised training.}
    \label{loss_plot}
    \end{figure}
    
    We have also discovered that fine-tuning the model with effective batch size of 512 after training with batch size 32 improves the performance significantly, especially when combined with selecting the well-performing networks by automatically running small number of games against the AIs.
\end{subsection}

\begin{subsection}{Reinforcement learning}
    
    We collect trajectories of length 32 using 24 environments, which send their inputs to a single thread that runs the network on batches of observations. Each time 512 trajectories are collected, an optimization step is done. We emulate using batch of size 512 by accumulating gradients during several iterations with batch of size 16, and doing the update optimizer step afterwards on the accumulated gradients.
    
    As mentioned earlier, we use Proximal Policy Optimization \cite{PPO}, as it is easy to implement and known for its stability. For value estimation, we add a single MLP layer to the network that takes the \textit{lstm\_output} as an input and predicts the value before the action is chosen.
    
    During reinforcement learning, we process around 500000 trajectories or 16M observation-action pairs. This is relatively low for reinforcement learning, especially with such a large action space, but due to computational limitations it is not possible to process many more iterations. 
    
    We train the agents against the built-in bots. The reward that the agent receives comes from several factors:
    \begin{itemize}
        \item Game result: 1 for the win, 0 for the draw, -1 for the loss;
        \item Resource cost of enemy units killed multiplied by 0.00003;
        \item Resource cost of enemy buildings destroyed multiplied by 0.0001;
        \item Collected resources - 0.00001 for each unit of minerals collected and 0.00003 for each unit of vespene collected.
    \end{itemize}
    
    The coefficients for the reward components were chosen to make the total reward in the same order of magnitude for different sources, while also keeping the game result as the dominating factor. It should be also noted that we use different value estimator outputs for the different reward components, and we sum their output to form a total prediction of the value.
        
    To further stabilize the training, we experimented with 3 different approaches:
    \begin{enumerate}
        \item Training the network using both reinforcement learning and supervised learning updates at the same time, so that it both improves through reinforcement learning and keeps correctly predicting player actions when observing a human replay.
        \item Applying the KL divergence penalty between the trained model and the original supervised model, with the supervised model's predictions computed on the same batch as the currently trained model, including the LSTM state stored at the beginning of the sequence (which means it experiences the LSTM states generated by the new model, which might shift significantly during training).
        \item Applying the KL divergence penalty between the trained model and the original supervised model, with the supervised model's predictions computed during trajectory generation (running the supervised network and storing its output each time the RL network runs). That way, the LSTM state for the supervised network is processed in the same way it would be processed if the network was actually playing the game.
    \end{enumerate}
    
    The approach 1 used $\gamma = 0.995$, approaches 2 and 3 used $\gamma = 0.999$.  
    
    We have discovered that the approach 1 was the least stable, and it only trained properly against Easy AIs. However, presumably because of the lack of KL divergence penalty, it has developed the most unusual strategy of the three. 
    
    \begin{figure}[th]
    \centering
    \includegraphics[width=12cm]{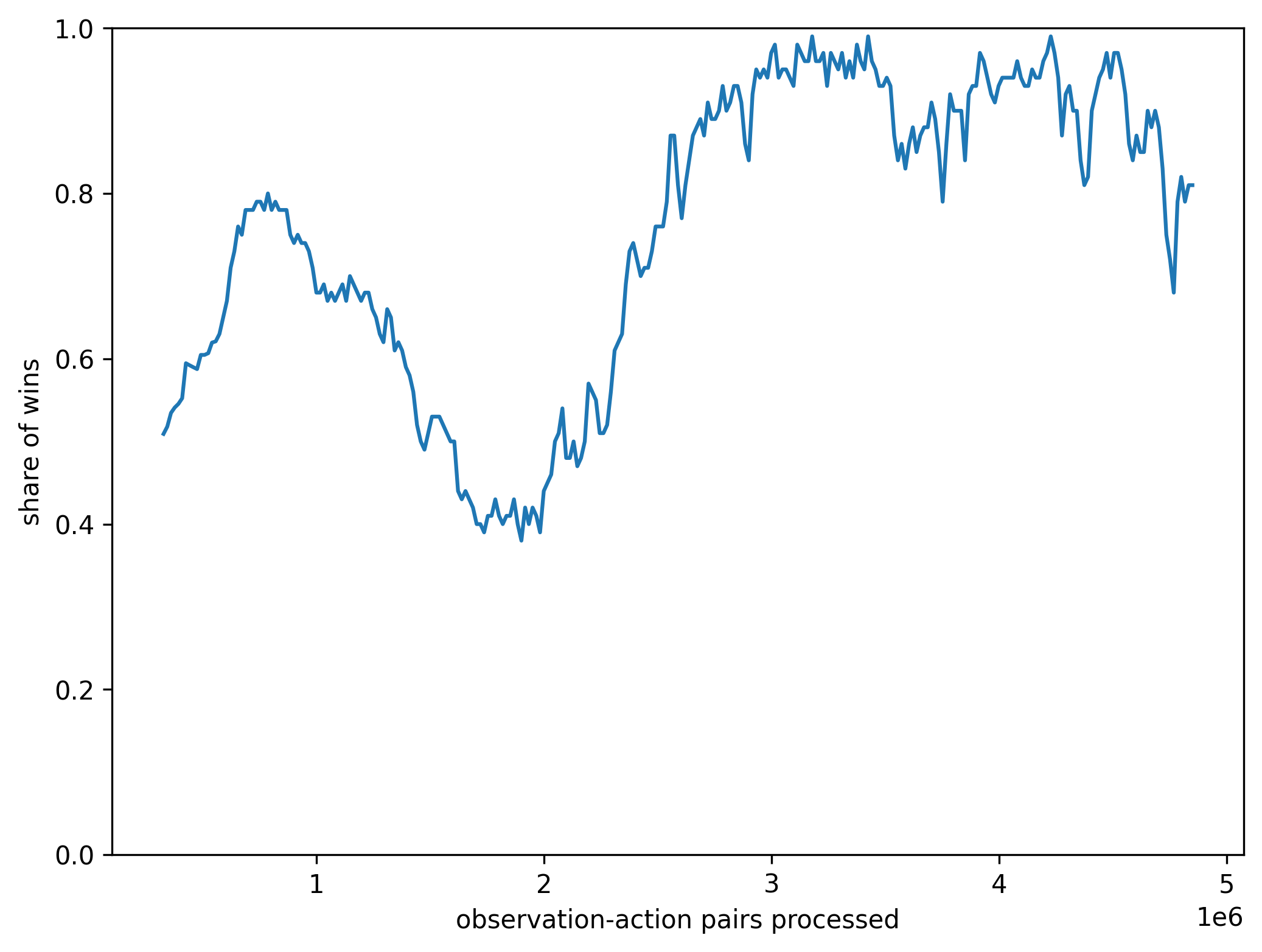}
    \caption{Win rate in the last 100 finished games (against the Easy Zerg AI) during reinforcement learning in approach 1.}
    \end{figure}
    
    Approaches 2 and 3 were successfully trained against the mix of Easy and Medium AIs. The approach 3 seemed like it was the most consistent and has achieved the highest performance against the mix, with the least significant drops of performance during training.  
    
    \begin{figure}[th]
    \centering
    \includegraphics[width=12cm]{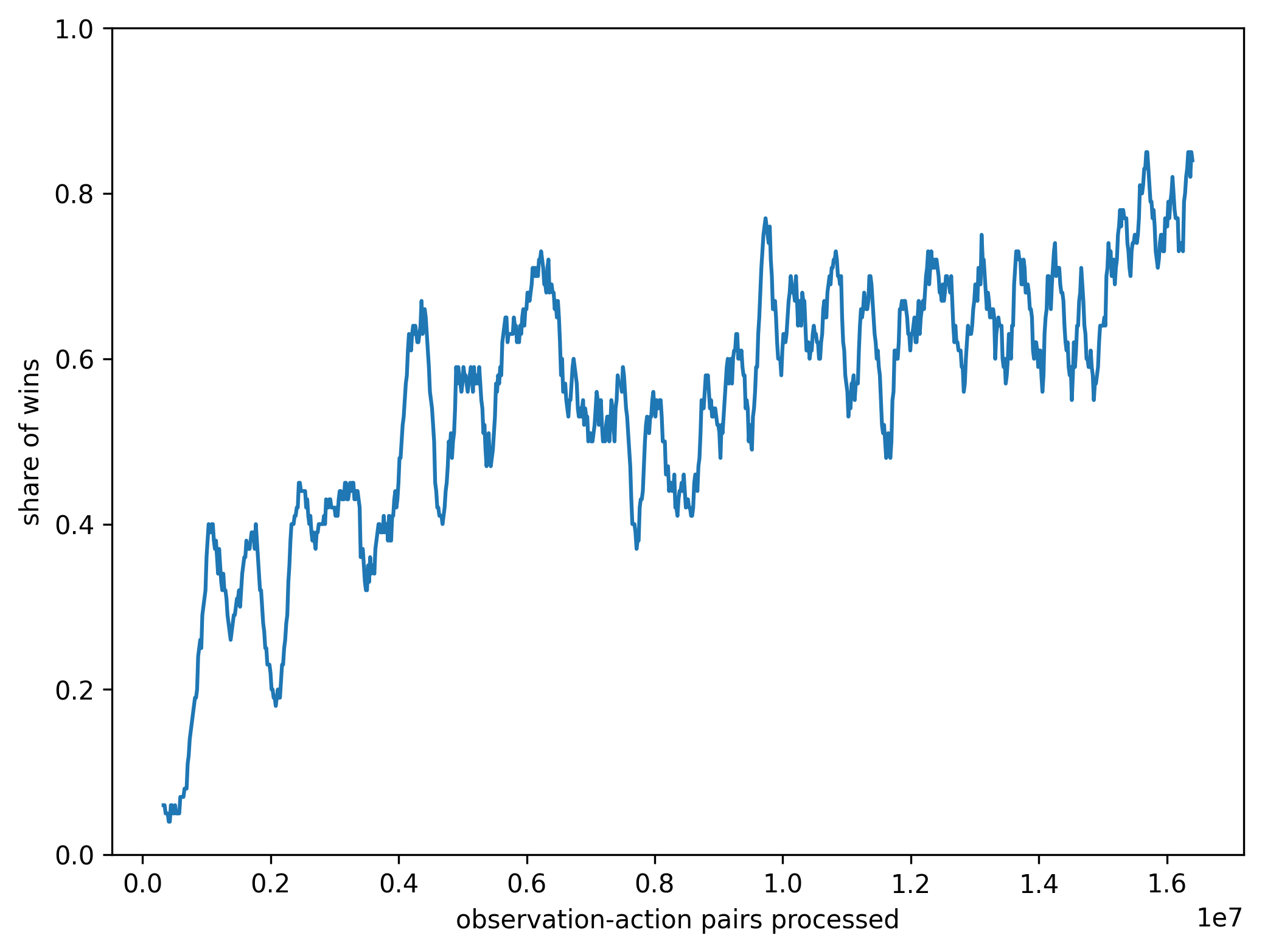}
    \caption{Win rate in the last 100 finished games (against the mix of Easy and Medium AIs) during reinforcement learning in approach 2.}
    \end{figure}
    
    \begin{figure}[th]
    \centering
    \includegraphics[width=12cm]{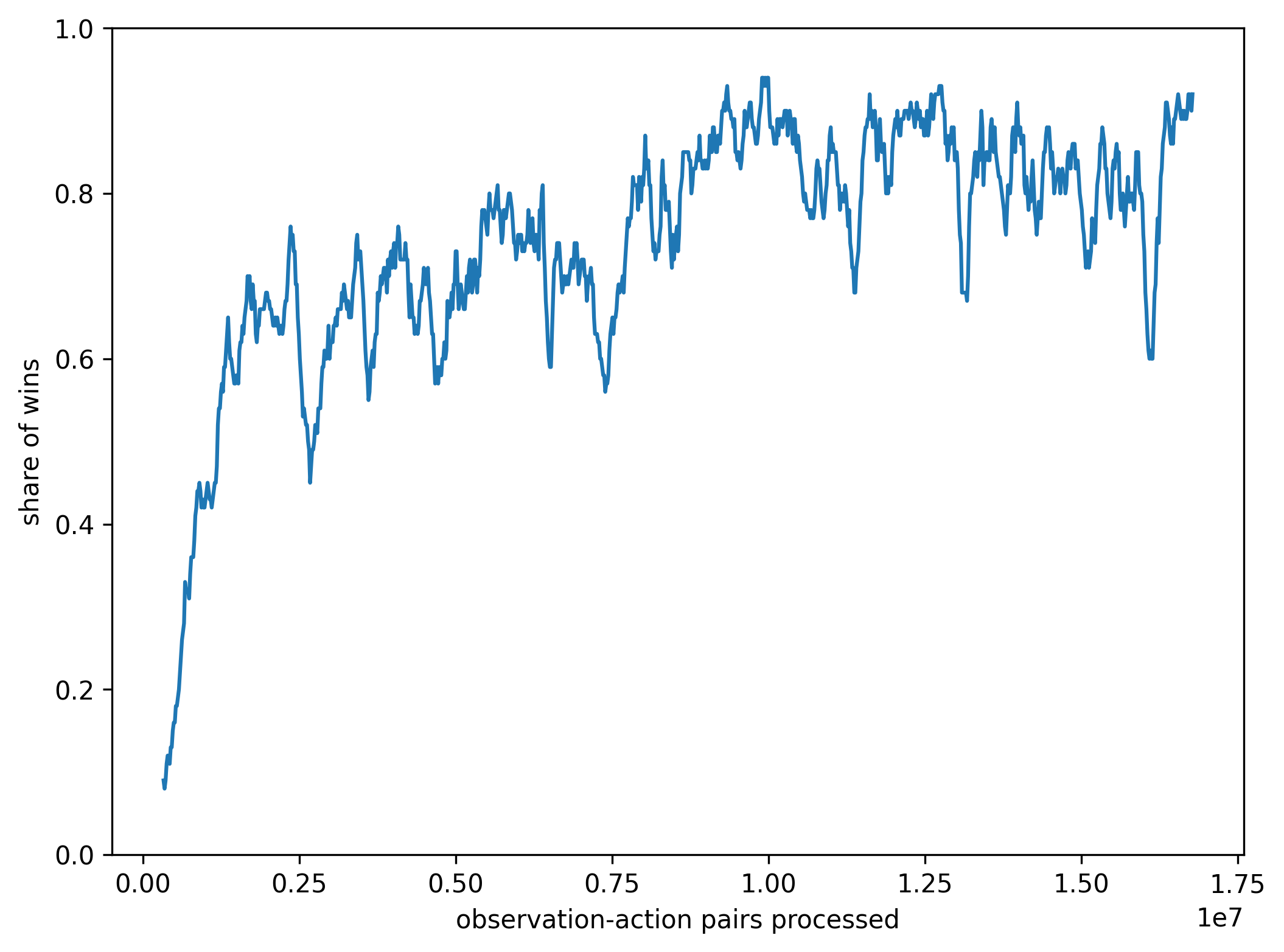}
    \caption{Win rate in the last 100 finished games (against the mix of Easy and Medium AIs) during reinforcement learning in approach 3.}
    \end{figure}
    
\end{subsection}

\chapter{Results}

\section{Performance}

To measure the performance of the agent we run multiple games against built-in scripted bots. We run the experiments using Very Easy, Easy, Medium and Hard AIs playing all three races, with our agent always playing as a Protoss.

For computational reasons, the games are run for 60 in-game minutes, and if the game does not finish in that time, we consider it a loss for our agent. Similarly, we treat true draws as losses.

\begin{table}[ht]
\centering
\begin{tabular}{|l|l|l|l|}
\hline
Bot Race \textbackslash Difficulty & Very Easy & Easy & Medium \\ \hline
Protoss                            & 49        & 25   & 3      \\ \hline
Terran                             & 55        & 27   & 3      \\ \hline
Zerg                               & 69        & 22   & 7      \\ \hline
\end{tabular}
\caption{Number of wins of the supervised agent (out of 100 games) against various in-game bots - the model saved after $\sim3$ days of training.}
\end{table}

\begin{table}[ht]
\centering
\begin{tabular}{|l|l|l|l|l|}
\hline
Bot Race \textbackslash Difficulty & Very Easy & Easy & Medium & Hard \\ \hline
Protoss                            & 88        & 63   & 18 & 0    \\ \hline
Terran                             & 91        & 61   & 17 & 0 \\ \hline
Zerg                               & 95        & 66   & 25 & 5    \\ \hline
\end{tabular}
\caption{Number of wins of the supervised learning agent fine-tuned on 512 batch size, with the best model automatically selected by running experiments consisting of 20 vs Easy and 20 vs Medium games during training.}
\end{table}

For the RL models, we have analyzed the performance of the 3 regularization approaches. All 3 approaches significantly improved the performance of the network trained with supervised learning. 

For the approach 1, the best result we got was when the agent was trained solely against Easy Zerg AI and discovered a very simple strategy consisting of building several 'Gateways' (basic Protoss infantry production buildings), producing around ten 'Zealots' (basic Protoss offensive units) relatively fast, and attacking the enemy with small waves of around 10 units each as fast as possible. Such strategies are usually called Rush. 

Networks in the approaches 2 and 3 were trained against a mix of Easy and Medium AIs, and it seems their strategies are quite similar to each other - they both start the game with a large number of Gateways, but appear to build a significant number of workers and bases, and they also build Stargates, which produce flying units. They also build quite a large number of defensive buildings in their bases, especially when attacked. 

\begin{table}[ht]
\centering
\begin{tabular}{|l|l|l|l|l|}
\hline
Bot Race \textbackslash Difficulty & Very Easy & Easy & Medium & Hard \\ \hline
Protoss                            & 100        & 91 & 21 & 3    \\ \hline
Terran                             & 100        & 99   & 83 & 17 \\ \hline
Zerg                               & 100        & 96   & 47 & 24    \\ \hline
\end{tabular}
\caption{Number of wins of the reinforcement learning agent from approach 1. The strategy it discovered appears to be least effective in mirror matchups, and very effective specifically against Terrans - we suspect this is because Terrans have no melee units, and AI is not capable of correctly controlling their ranged units for maximum effectiveness.}
\end{table}

\begin{table}[ht]
\centering
\begin{tabular}{|l|l|l|l|l|}
\hline
Bot Race \textbackslash Difficulty & Very Easy & Easy & Medium & Hard \\ \hline
Protoss                            & 99        & 93 & 39 & 4    \\ \hline
Terran                             & 99        & 98   & 68 & 13 \\ \hline
Zerg                               & 100        & 95   & 61 & 11    \\ \hline
\end{tabular}
\caption{Number of wins of the reinforcement learning agent from approach 2. Mirror matchups still remain the hardest for this network, but the differences in effectiveness against different races appear smaller.}
\end{table}

\begin{table}[ht]
\centering
\begin{tabular}{|l|l|l|l|l|}
\hline
Bot Race \textbackslash Difficulty & Very Easy & Easy & Medium & Hard \\ \hline
Protoss                            & 100        & 98   & 54 & 1    \\ \hline
Terran                             & 98        & 98   & 72 & 6 \\ \hline
Zerg                               & 100        & 98   & 66 & 5    \\ \hline
\end{tabular}
\caption{Number of wins of the reinforcement learning agent from approach 3. It appears the performance against the Medium AI has improved, especially for mirror matchups, but the network did not perform as well against the Hard AI as the previous ones.}
\end{table}

\section{Conclusion}

We have presented a neural network architecture and a training pipeline that can produce agents for playing StarCraft II using limited computational resources. Unlike other small-scale approaches for playing the full game, we use a single neural network for the entirety of decision-making.

By evaluating our agents in a large number of two-player games, we have shown that our approach is capable of training agents that can compete with some of the built-in bots. Notably, all 3 RL experiments resulted in significant improvement over the original supervised agents.  

Some of our agents were also surprisingly effective against more difficult bots after only training against a fixed, easy opponent - an agent trained only against the Easy Zerg AI developed a strategy that was moderately effective even against the Hard AI. 

\section{Future works}

Even with our relatively small computational requirements, our agents still take multiple days to train, which significantly limited the amount of experiments we were able to perform. In the future, we hope to explore the following topics:
\begin{itemize}
    \item Removing the limitation of playing the game on a single map;
    \item Exploring different neural network architectures for faster training and better performance;
    \item Using newer reinforcement learning algorithms;
    \item Introducing learning through self-play and a league of agents as proposed in AlphaStar.
\end{itemize}

\bibliographystyle{plain}
\bibliography{references}

\end{document}